\documentclass[12pt]{article}

\usepackage{amsmath,amssymb,amsthm}
\usepackage{newtxtext}
\usepackage{newtxmath}

\usepackage{enumitem}
\setlist{leftmargin=5.5mm}

\usepackage{enumitem}
\usepackage{pdfpages}
\usepackage[all,curve,2cell]{xypic}
\UseAllTwocells
\newtheorem{thm}{Theorem}[section]
\newtheorem{defn}[thm]{Definition}
 \usepackage[implicit,bookmarks,colorlinks=true,linkcolor=blue]{hyperref}

 \def\C{\mathcal{C}}
 \def\I{\mathcal{I}}

\begin{document}

\title{In between myth and reality: AI for math\\
\Large -- a case study in category theory --}
\author{R\u{a}zvan Diaconescu
\thanks{Simion Stoilow Institute of Mathematics of the Romanian
  Academy, email: razvan.diaconescu@ymail.com}}

%
% Journal of Mathematics and Artificial Intelligence (JMAI) ?
% 

\maketitle

\begin{abstract}
\noindent
Recently, there is an increasing interest in understanding the
performance of AI systems in solving math problems.
A multitude of tests have been performed, with mixed conclusions. 
In this paper we discuss an experiment we have made in the direction
of mathematical research, with two of the most prominent contemporary
AI systems.
One of the objective of this experiment is to get an understanding of
how AI systems can assist mathematical research.
Another objective is to support the AI systems developers by
formulating suggestions for directions of improvement.
\end{abstract}

%%% ----------------------------------------------------------------------
\maketitle
%%% ----------------------------------------------------------------------
% \tableofcontents

\section{Introduction}

Unsurprisingly, mathematics is one of the main benchmarks for current
AI.
There are even significant efforts to build AI systems dedicated to
mathematics.
One of them is o3-mini \cite{o3-mini}, developed by OpenAI.
It is claimed that it solved 80\% of the subject sheet at American
Invitational Mathematics Examination (AIME) 2024 \cite{openAI2024c}, a 
prestigious competition leading to the USA Mathematical Olympiad.  
Another one is Grok-3 \cite{grok-3}, developed by xAI (an Elon Musk
company), which is also claimed to be very good at math and physics.
These claims are in stark contrast with the statistics put forward in 
\cite{frontiermath} where, in the case of mathematical research, the
AI can solve only 2\% of the problems.   

Our motivation for this AI experiment was to try to understand, from
the perspective of a non-specialist in machine learning, what can this
kind of AI do for the working mathematicians, how can we use it to
support our work, and what is behind the vast gap (claimed in
\cite{frontiermath}) between what current AI capabilities and the
prowess of the mathematical research community. 
% We did not expect it to work perfectly and we were ready to ‘extract’
% anything useful from its output. 

The readership target of this paper consists mainly of the
mathematicians with a moderate degree of fluency with elementary
category-theoretic thinking.

\subsection*{On methodology}

The methodology for our experiment had three components:
\begin{enumerate}

\item The choice of an `adequate' mathematical area.
  For this, we envisaged the following criteria:
  \begin{itemize}[leftmargin=1em, itemsep=-2pt]

  \item Should be a well-recognised one, with a vast literature.
    The AI systems should access easily the basic sources. 

  \item Should have a high level of standardisation.
   This helps avoiding data ambiguity and confusion. 

 \item We have a high level of expertise in the area.
   Thus, we can have a correct asesment of the performance of the AI. 

  \end{itemize}
  Under these criteria, category theory \cite{maclane98} arose as a
  natural choice.   

\item The choice of an `adequate' problem.
  The problem should fullfil the following requirements:
  \begin{itemize}[leftmargin=1em, itemsep=-2pt]

\item It should be easy to formulate in a simple and clear
  way.

\item It should have a relatively straightforward solution, being
  of the nature of an exercise in a graduate course rather than a
  research problem. 

\item It should not belong to the corpus of well-known exercises that
  we can easily find in textbooks.
  For that, we thought of involving a simple categorical concept that,
  on the one hand is not that standard, but on the other hand appears
  in a significant number of books or articles.
  Such a concept is that of \emph{inclusion system}.
  The paper introducing inclusion systems has now over 300 citations
  on Google Scholar, from which we estimate that at least two thirds
  refer to inclusion systems.

\end{itemize}
  One may think that a choice of a topic and of a problem that are
  less mainstream may be a deliberate hindrance to the AI systems.
  However, we have to consider that these are precisely the kind of
  problems that professional researchers in mathematics engage with,
  rather than those that have already been intensely studied. 

\item In analysing the results obtained with the AI, we looked for the
  following aspects:
  \begin{itemize}[leftmargin=1em, itemsep=-2pt]

  \item The result of data gathering. 

  \item The language of mathematics.

  \item The reasoning. 

  \end{itemize}
  We consider all above aspects as being part of the solution. 
  
\end{enumerate}
The structure of the paper is as follows.
First we introduce the topic and the problem, and present its solution.
Then we provide an analysis of the solutions given by the two AI
systems.
Finally, we daw some conclusions and formulate some recommendations,
for the users, and for the developers. 

\section{The problem and its solution}

The benchmark problem is an exercise within the theory of inclusion
systems.
It is about a property that is somehow part of the folklore of the
area, and  appears even as an exercise in the monograph \cite{iimt2},
but, up to our knowledge, does not have a solution published anywhere. 
In this section we first present very briefly the concept of inclusion
system, and then we formulate the benchmark problem and present a
solution in the way it is usually done by human mathematicians. 

\subsection{Inclusions systems}

Inclusion systems were introduced in theoretical computer science studies
\cite{modalg} but in the meantime it has been used quite intensively
in both computer science and logic (especially model theory).
This means there is a relatively large literature either developing
inclusion systems or using them, the latter being prevalent. 
Let us here recall its definition. 
\begin{defn}%[Inclusion systems]\label{inc-sys-dfn}
  A pair of categories $(\I, {\mathcal{E}})$ is an \emph{inclusion
    system} for a category $\C$ if $\I$ and $\mathcal{E}$ are
  two broad subcategories of $\C$ (i.e. $|\I| = | \mathcal{E}| =
  |\C|$) such that 
  \begin{enumerate}[leftmargin=1.5em, itemsep=-2pt] 

  \item $\I$ is a partial order (with the order relation denoted by
    $\subseteq$), and

  \item every arrow $f$ in $\C$ can be factored uniquely as a
    composition $f = e_f ; i_f$ (written in diagrammatic order) with
    $e_f \in \mathcal{E}$ and $i_f \in \I$.
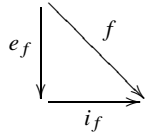
    \[
      \xymatrix @+2ex {
        \ar[dr]^f \ar[d]_{e_f} & \\
        \ar[r]_{i_f} &
        }
    \]
    
  \end{enumerate}
  The arrows of $\I$ are called \emph{abstract inclusions}, and  
  the arrows of $\mathcal{E}$ are called \emph{abstract surjections}.
\end{defn}
This definition is standard since many years, though it updates
slightly the original one from \cite{modalg}.
Although this may be irrelevant for this experiment, it is may be
worth saying that inclusion system represent an abstract axiomatic
approach to the notions of substructures (given by $\I$) and quotient
structures (given by $\mathcal{E}$).
This applies to myriad categories, such as categories of various
species of algebras, of topological spaces, of models of various
logical systems, to syntactic structures in computer science, etc.
Relevant examples of inclusion systems are countless. 

\subsection{The problem}
\label{prob-sec}

We chose the following exercise from \cite{iimt2}:
\begin{quote}
Prove that in any category with an inclusion system each cospan (sink)
of abstract inclusions that has a pullback, has a unique pullback 
consisting of abstract inclusions.
\end{quote}
The proof consists of the following steps:
\begin{itemize}[leftmargin=1.5em, itemsep=-2pt]

\item Consider a cospan of abstract inclusions $(i_1, i_2)$ and a
  pullback cone $(f_1, f_2)$ for that.
  
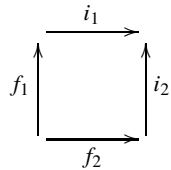
\[
    \xymatrix @+2ex {
      \ar[r]^{i_1}  & \\
      \ar[u]^{f_1} \ar[r]_{f_2} & \ar[u]_{i_2}
    }
  \]
  We will try to get a pullback cone of abstract inclusions from
  $(f_1, f_2)$, and then to prove its uniqueness.

\item For that, we factor $f_1 = e_1 ; i'_1$ through the inclusion
  system.
  
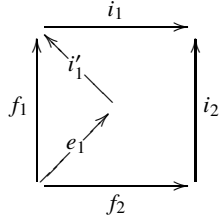
\[
    \xymatrix {
      \ar[rr]^{i_1}  &  & \\
      & \ar[ul]|{i'_1} & \\
      \ar[uu]^{f_1} \ar[rr]_{f_2} \ar[ur]|{e_1} & & \ar[uu]_{i_2}
      }
  \]

\item By the \emph{Diagonal Fill-in} property of the inclusion systems
  (see \cite{iimt2}) we get an unique $h$ such that $e_1 ; h = f_2$
  and $i'_1 ; i_1 = h ; i_2$.
  
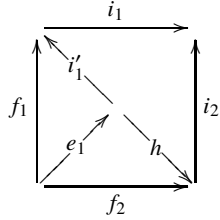
\[
    \xymatrix  {
      \ar[rr]^{i_1}  &  & \\
      & \ar[ul]|{i'_1} \ar[rd]|h& \\
      \ar[uu]^{f_1} \ar[rr]_{f_2} \ar[ur]|{e_1} & & \ar[uu]_{i_2}
      }
  \]

\item Then by factoring $h$ through the inclusion system and by the
  uniqueness of the factorisations, we get that $h$ is an abstract
  inclusion.

\item Now, we prove that $(i'_1, h)$ is a pullback cone for the cospan
  $(i_1, i_2)$.
  Let $(g_1, g_2)$ be a cone for that cospan.
  
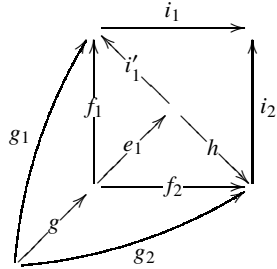
\[
    \xymatrix {
       & \ar[rr]^{i_1}  &  & \\
       & & \ar[ul]|{i'_1} \ar[rd]|h& \\
       & \ar[uu]|{f_1} \ar[rr]|{f_2} \ar[ur]|{e_1} & & \ar[uu]_{i_2} \\
       \ar[ur]|g \ar@/^/[uuur]^{g_1} \ar@/_/[urrr]_{g_2} & & & 
      }
  \]
  \begin{itemize}[leftmargin=1.5em]

  \item From the pullback property of the cone $(f_1, f_2)$, let $g$
    be the unique arrow such that $g; f_i = g_i$, $i=1,2$.

  \item Then $(g; e_1); i'_1 = g_1$ and $(g;e_1);h = g_2$.
    Moreover, $g;e_1$ is unique with these properties by relying on
    the general mono property of abstract inclusions applied to $i'_1$
    and $h$.\footnote{This mono property is a well-known consequence
      of the axioms of inclusion systems.}

  \end{itemize}

\item Finally, we have to prove that this pullback cone of abstract
  inclusions is unique.
  For that, we assume another such pullback, and by the universal
  property of the pullbacks and by the uniqueness of the
  factorisations we get two mediating arrows between these two
  pullback cones.
  These are abstract inclusions which are inverse to each other,
  which means that they are identities. 
  
\end{itemize}

\section{An analysis of the proofs provided by the AI systems}

We analyse them on the three aspects mentioned in the introduction to
this paper.
We mention that specific prompts and repeated trials did not make any
significant diference with respect to what solutions we got from both
AI systems. 

\subsection{The data}

Gathering relevant data is critical for solving anything.
In the case of our problem, the single most important data is the
definition of the concept of inclusion system.
There is a significantly large literature where inclusion systems are
defined, including several dozens of published articles, not
mentioning the monograph \cite{iimt2} (its first edition being
published already in 2008).
However, to our surprise, both AI systems seemed to have problems with
this, quite catastrophically in the case of Grok.
\begin{itemize}[leftmargin=1.5em, itemsep=-2pt]

\item o3-mini seemed to be quite succesful with this, although the way
  it reminds us this definition is quite clumsy.
  For someone without conventional access to that definition, it would
  be impossible to recover it from the o3-mini output.
  But someone who is already familiar with inclusion systems, would
  recognise that definition as somehow correct.
  Having said that, the o3-mini concept of inclusion system relies on
  that of `factorisation system' which is not discussed.
  The problem here is that the category theory literature contains
  several slightly different concepts of `factorisation systems'.
  If there was at least a reference for this, then we would know which
  one to consider, but o3-mini did not provide this.
  And in mathematics `slight difference' may mean in fact huge
  difference.
  So, many important parts from the definition of inclusion systems
  were missing, and we had to assume them from a presumed proper
  concept of factorisation system.

\item Grok failed badly with the definition of inclusion system,
  as it came up with a completely messed up concept, with crucial
  parts missing, and instead a derived property (i.e. the `stability
  under pullbacks') being made part of the definition. 
  Incidentally, that property plays an important role in the solution
  to the proposed problem.
  However, even that was formulated incorrectly, confusing between
  `exists' and `any'. 
  It is much more case that Grok presented a collection of properties
  that it was going to use in the proof, than the actual concept. 
  By reading the output provided by Grok, and by comparing it with our
  definition, the reader may understand well by himself the amplitude
  of this critical failure. 
  
\end{itemize}

\subsection{Language of mathematics}

Common mathematical language, including the way the mathematicians
write things, has certain clear features.
In their absence we may easily feel that we are not in the situation
of communication of mathematics.
One such feature is precision.
Both o3-mini and Grok display some very strange imprecise mathematical
language, something that is considered by mathematicians, if not
unacceptable then certainly annoying.
For instance: ``every arrow factors \emph{essentially} uniquely'',
``must be (\emph{uniquely isomorphic to}) the $I$-part'',
``\emph{in many formulations one can prove (or assume)}'', ``a
composite of an $I$–morphism with an $E$–morphism has an
\emph{$I$–factorization}'', ``there is no \emph{extra “noise” coming
  from the $E$–part}'' (o3-mini) or ``an inclusion system is
\emph{typically} a class...'' (Grok), etc. 

But o3-mini failed also in a completely unacceptable way with the
mathematical language, this time at the level of mathematical
formulas.
Writing non-sensical mathematical formulas may be one of the most
grave failures when communicating mathematics.
What happened is the following.
In category theory the notation for arrow composition can occur in two
different ways.
\begin{itemize}[leftmargin=1.5em, itemsep=-2pt]

  \item On the one hand, there is set-theoretic way, which is used by
    most `pure' mathematicians.
    This notation comes from the way we usually write the composition
    of functions.
    If $f \colon A \to B$ and $g \colon B \to C$ then their composition is
    denoted by $g \circ f$.
    There are two aspects of this notation: the use of the symbol
    $\circ$ and the order in which $f$ and $g$ appear.
    These two are usually inter-dependent.
    That people denote abstract arrow composition like that comes from
    their strong dependence on the set theory language, which pervades
    almost all mathematical practice.

\item On the other hand, there is the way the computer scientists
  denote arrow composition, which is the `diagrammatic' way.
  This means that the in the composition, the two arrows appear in the
  order we see them when drawing a diagram.
  \[
    \xymatrix{
      A \ar@/_/@{.>}[rr]_{f;g} \ar[r]^f & B \ar[r]^g & C 
      }
  \]
  Naturally, the symbol used for composition is $;$ rather than
  $\circ$.
  Apparently, this has to do with program composition, as in many
  programming the symbol $;$ serves as a separator that serves also to
  link pieces of programs together.
  And programs have input and output (in fact like functions), so they
  can be naturally thought as arrows.
  This is how categorical semantics treats programs.
  Anyway, for people that involve complicated diagrams in their work,
  the computer science notation for arrow composition is more
  intuitive. 

\end{itemize}
Then o3-mini, within a single equality used both kinds of composition
but in a maximally confusing way, using always the same symbol
($\circ$) but with the two diferent orders discussed above.
For instance in $p = e_p \circ i_p$ should have been written correctly
either as $p = i_p \circ e_p$ or as $p = e_p ; i_p$.
A consequence of this is that the formulas \eqref{o3-1} and
\eqref{o3-2} do not parse (see Appendix \ref{o3-mini-appendix}).
In the case of \eqref{o3-3}, the left-hand side is written correctly
while the right-hand side is written incorrectly. 
%Many other times $\circ$ is used correctly. 

Another failure of o3-mini is represented by the formulas \eqref{o3-4},
which, strictly speaking, are non-sensical, although one may guess
what these try to convey.

\subsection{The `reasoning'}

Now let us analyse the reasoning of these AI systems. 

\paragraph{o3-mini.}
Modulo the numerous awkward formulations and messed up
formulas (related to the confusion about composition discussed above), 
we can say that o3-mini did produce a valid proof, more or less along
the same lines as our proof from Section \ref{prob-sec}.
The only difference is that o3-mini factored the diagonal of the
starting pullback square rather than one of the sides of the pullback
cone.
Consequently, it did not have to rely on the \emph{Diagonal Fill-in
  Lemma}.
This evaluation applies to the first half of the proof, until the
square of inclusions is obtained.
The last half of the proof, which is non-trivial, namely to prove that
the square of inclusions is a pullback indeed, and that it is unique
(these corresponding to the last two steps of our proof), were
expedited in a couple of phrases that provide just very general hints
about how these can be done. 

\paragraph{Grok-3.}
The proof provided by Grok was compromised from the start due to the
wrong definition for inclusion systems.
However, we can analyse the correcteness of the its reasoning based on
the hypotheses it wrongly `believes' to define the concept of inclusion
system.
Under these circumstances, Grok developed a very simple argument
about the existence of an inclusive pullback of inclusions, just by
invoking twice the `stability under pullbacks'.
For the uniqueness part it came up with a pretty twisted argument,
that was also based on a heavy misunderstanding of inclusion systems.

\section{Conclusions and recommendations}

We have selected a problem that was balanced in the sense that the AI 
system was supposed to have a relatively easy access to the relevant
literature, which albeit not abundant was still reasonable large.
The problem had a clean straightforward solution, but not a trivial
one which also, up to our knowledge, is not available as such in the
literature.
In brief, we have found the following facts:
\begin{itemize}[itemsep=-2pt]

\item In general, the o3-mini and the Grok system behaved highly 
  differently. 

\item Surprisingly, both AI systems had difficulties with gathering
  the relevant data, namely the concept of inclusion system.
  That was mild in the case of o3-mini but catastrophic in the case of
  Grok.
  The conclusion here is that we should never rely on retrieving
  mathematical concepts and definitions through AI systems.
  We should just use books, articles, etc., and we have to do this
  with a human mind.
  Trying to understand this issue, we know that in the inclusion
  systems literature, often (but not always) it is the case that the
  definition of inclusion systems is recalled without writing it
  explicitly in a definition environment. 
  Such environments are often reserved for new definitions that are
  introduced by the respective publication.
  It seems that, from a quite narrative text, but using precise and
  explicit mathematical language, the current AI technology is not
  capable to `understanding' what is a definition and what is not.
  It has difficulty to extract the necessary mathematical concepts. 
  
\item In both cases, the systems used an inappropriate language of
  mathematics.
  Moreover, o3-mini used repeatedly some incoherent mathematical
  notations. 

\item While o3-mini performed some non-trivial valid reasoning,
  basically solving the first half of the problem (the existence
  part), Grok performed only some almost trivial reasoning.
  However, in the case of Grok, this experiment was heavily
  compromised by the bad failure in gathering the concept of inclusion
  system.
  We can say, that in the case of someone who wanted to get a
  solution for this problem without doing it by himself, modulo the
  language and notational problems, the o3-mini provided a valid
  solution for the first part.
  Concerning the second part of the problem (i.e. the uniqueness
  part), both outputs were relatively useless.
  While o3-mini expedited it in a couple of phrases, without the
  necessary details, Grok remained the prisoner of its own failure to
  gather the appropriate concept of inclusion system. 
  
\end{itemize}

We should also mention that we have also tried another related
experiment, with o3-mini only, this time with another exercise from
\cite{iimt2}, significantly harder and involving more concepts.
In that case there was nothing to analyse because we got a complete
hallucination.

From this experience we can draw the following recommendations:
\begin{itemize}

\item For developers of soft AI systems doing math: it is important to
  build some meta-level layer that checks the coherence of the data,
  of the mathematical notations, and detects and corrects imprecisions
  in the mathematical language. 

\item For users (working mathematicians): do not rely on the data
  provided by the AI systems (such as definitions of concepts or even
  stated results) but rather check them with the relevant literature.
  Do not have unrealistic expectations.
  Sometimes, the AI system can aggregate useful ideas (like 03-mini
  did in its `reasoning'), but there cannot be any proper
  understanding-based reasoning underlying that, because any non-human
  system is bound to hit the wall of what we call `understanding',
  which is a feature of the human mind that fundamentally cannot even
  be defined. 
  The understanding aspect is also very related to concept creation,
  as advanced mathematics relies heavily of sophisticated conceptual
  constructions, something that cannot be expected from any form of
  AI. 
  These inherent limitation have been explained thoroughly in
  \cite{penrose-emperor}.

\item Again for the users: when looking for AI assistance with your
  research problems try to combine soft AI assistance with hard AI
  alternatives, such as theorem provers.
  Automatic theorem provers such as Vampire
  \cite{vampire}, Spass \cite{spass3.5} or E
  \cite{Schulz-Cruanes-Vukmirovic:E-2.3-2019} may at least formally
  verify some results that can be encoded in a first-order logic
  form.   

\end{itemize}

%\subsection*{Acknowledgment}

% \bibliographystyle{spbasic}      % basic style, author-year citations
\bibliographystyle{spmpsci}        % mathematics and physical sciences
\bibliography{/users/diacon/TEX/tex.bib}   % name your BibTeX data base

\appendix
\section{The o3-mini output}
\label{o3-mini-appendix}

% \includepdf[scale=0.7,pages=-]{ai&inclusionsystems.pdf}

Let $\C$ be a category equipped with an inclusion system—that is, a
factorization system $(E,I)$ where the “inclusion‐morphisms” in $I$
are thought of as “abstract inclusions” (for example, there is at most
one arrow in $I$ between any two objects, so that $I$ gives a partial
order on subobjects), and every arrow factors essentially uniquely as
an $E$–morphism followed by an $I$–morphism.  

Now suppose we are given a cospan of abstract inclusions

\[
A \overset{i}{\longrightarrow} C \overset{j}{\longleftarrow} B
\]

with \(i,j\in I\), and suppose that the pullback of \(i\) and \(j\)
exists. (A “pullback” here is the limit of the diagram \(A
\overset{i}{\to} C \overset{j}{\leftarrow} B\).) 

Let
\[
\xymatrix{
P \ar[r]^{p} \ar[d]_{q} & A \ar[d]^{i} \\
B \ar[r]_{j} & C
}
\]
be a pullback cone; that is, \(i\circ p=j\circ q\) and it is universal
among such cones. 

Because our category has an inclusion system, each morphism factors
uniquely as an $E$–morphism followed by an $I$–morphism. In particular,
factor the projections \(p\) and \(q\) as 
\[
p = e_p \circ i_p\quad\text{and}\quad q = e_q \circ i_q,
\]
where
\[
e_p\in E,\quad i_p\in I,\qquad \text{and}\qquad e_q\in E,\quad i_q\in I.
\]

We now show that, after “improving” the pullback cone by replacing
\(p\) and \(q\) with \(i_p\) and \(i_q\) (which lie in $I$), the
resulting cone is still a pullback and is unique among cones whose
legs are abstract inclusions. 

\paragraph{Step 1. Comparing the Two Factorizations.}
Since \(i\) is in $I$ and $I$–morphisms are “rigid” (they are monic and
essentially unique), consider the composite 
\begin{equation}\label{o3-1}
i\circ p = i\circ (e_p\circ i_p) = (i\circ e_p)\circ i_p.
\end{equation}
Similarly, we have
\begin{equation}\label{o3-2}
j\circ q = j\circ (e_q\circ i_q) = (j\circ e_q)\circ i_q.
\end{equation}
But by the pullback property,
\[
i\circ p = j\circ q.
\]
Now factor the common composite \(i\circ p = j\circ q\) itself (using
the inclusion system) as 
\begin{equation}\label{o3-3}
i\circ p = e\circ i',
\end{equation}
with \(e\in E\) and \(i'\in I\).

The uniqueness of factorization in the inclusion system tells us that
the $I$–part \(i'\) of the composite must be (uniquely isomorphic to)
the $I$–part obtained from the two different factorizations. In
particular, the factorizations 
\[
(i\circ e_p)\circ i_p \quad\text{and}\quad (j\circ e_q)\circ i_q
\]
must have the same “$I$–part” (up to the unique isomorphism provided by
the factorization system). In many formulations one can prove (or
assume) that if a composite of an $I$–morphism with an $E$–morphism has an
$I$–factorization, then the $E$–component must be an
isomorphism. (Intuitively, once the “inclusion‐part” is fixed, there
is no extra “noise” coming from the $E$–part.) Therefore, it follows
that both \(e_p\) and \(e_q\) are isomorphisms. Consequently, the
original projections \(p\) and \(q\) are (up to isomorphism)
themselves in $I$; indeed, we have 
\begin{equation}\label{o3-4}
p\cong i_p \quad\text{and}\quad q\cong i_q,
\end{equation}
with \(i_p\) and \(i_q\) in $I$.

\paragraph{Step 2. The Improved Pullback Cone.}
Thus, replacing the original pullback projections \(p\) and \(q\) by
the (isomorphic) arrows \(i_p\) and \(i_q\) (which are in $I$) we obtain
a cone 
\[
\xymatrix{
P \ar[r]^{i_p} \ar[d]_{i_q} & A \ar[d]^{i} \\
B \ar[r]_{j} & C
}
\]
which still commutes (since isomorphic modifications of the legs do
not affect the commutativity) and by the universal property of the
pullback remains a pullback cone. Moreover, by the uniqueness part of
the factorization system, this “inclusion‐pullback” is unique up to
the unique isomorphism. 

\paragraph{Conclusion.}
We have proved that if a cospan of abstract inclusions\linebreak
\(A\overset{i}{\to} C\overset{j}{\leftarrow} B\) has a pullback in $\C$,
then one may (uniquely up to isomorphism) choose the pullback so that
its projection maps \(P\to A\) and \(P\to B\) belong to $I$—that is,
they are abstract inclusions. This completes the proof.

\section{The Grok-3 (think) output}

To prove that in any category with an inclusion system, each co-span
of abstract inclusions that has a pullback has a unique pullback
consisting of abstract inclusions, let’s proceed step-by-step with a
clear and formal argument. We’ll define the necessary concepts, state
the assumptions, and construct the proof using the properties of an
inclusion system. 

\paragraph{Define the Key Concepts.}
Consider a category $\C$ equipped with an \emph{inclusion system}.
An inclusion system is typically a class $I$ of monomorphisms in $\C$
satisfying certain properties that allow them to behave like
"inclusions" (e.g., embeddings or subobject inclusions in familiar
categories like sets or topological spaces). For the purposes of this
proof, we assume $I$ has the following standard properties: 
\begin{itemize}[itemsep=-2pt]
  
\item \emph{Closure under composition:} If \( f: A \to B \) and \( g: B \to C
  \) are in $I$, then \( g \circ f: A \to C \) is in $I$.
\item \emph{Stability under pullbacks:} In a pullback square
\[
\xymatrix{
  P \ar[r]^{p_2} \ar[d]_{p_1} & B \ar[d]^g \\
  A \ar[r]_f & C
}
\]
if \( g: B \to C \) is in $I$, then \( p_1: P \to A \) is in
$I$. Symmetrically, if \( f: A \to C \) is in $I$, then \( p_2: P \to
B \) is in $I$.
\end{itemize}

These properties ensure that inclusions behave consistently under
categorical constructions like pullbacks, which is central to our
proof. 

A \emph{co-span} in $C$ is a diagram of the form:

\[
A \xrightarrow{i} C \xleftarrow{j} B
\]

where \( i: A \to C \) and \( j: B \to C \) are morphisms. The problem
specifies that this co-span consists of \emph{abstract inclusions},
meaning \( i, j \in I \). We are given that this co-span \emph{has
a pullback}, and we need to show that it has a \emph{unique pullback
consisting of abstract inclusions}, i.e., a pullback where the
projection morphisms are in $I$. 

A \emph{pullback} of the co-span \( A \xrightarrow{i} C \xleftarrow{j}
B \) is an object \( P \) with morphisms \( p_1: P \to A \) and \(
p_2: P \to B \) such that: 
\begin{itemize}[itemsep=-2pt]
\item \( i \circ p_1 = j \circ p_2 \), and
\item The universal property holds: for any object \( Q \) with morphisms
\( q_1: Q \to A \) and \( q_2: Q \to B \) satisfying \( i \circ q_1 =
j \circ q_2 \), there exists a unique morphism \( u: Q \to P \) such
that \( p_1 \circ u = q_1 \) and \( p_2 \circ u = q_2 \). 
\end{itemize}
The pullback can be depicted as:
\[
\xymatrix{
  P \ar[r]^{p_2} \ar[d]_{p_1} & B \ar[d]^j \\
  A \ar[r]_i & C
}
\]

The phrase “consisting of abstract inclusions” means that \( p_1: P
\to A \) and \( p_2: P \to B \) are both in $I$. Since pullbacks in
category theory are unique up to unique isomorphism, “unique pullback”
typically refers to uniqueness up to isomorphism, and we need to
verify that the pullback has this property. 

\paragraph{Step 2: State the Goal.}
Given a co-span \( A \xrightarrow{i} C \xleftarrow{j} B \) where \( i,
j \in I \), and assuming a pullback exists, we must prove two
things: 
\begin{enumerate}[itemsep=-2pt]
  \item \emph{Existence}: There exists a pullback \( P \) with \( p_1:
    P \to A \) and \( p_2: P \to B \) both in $I$.
    \item \emph{Uniqueness}: This pullback is unique up to
      isomorphism, and any such pullback has its projection morphisms
      in $I$. 
\end{enumerate}
Since pullbacks are always unique up to isomorphism in a category
(when they exist), the key task is to show that the pullback of this
co-span has projection morphisms in $I$, leveraging the properties of
the inclusion system. 

\paragraph{Step 3: Construct the Proof.}
Assume the co-span \( A \xrightarrow{i} C \xleftarrow{j} B \) with \(
i, j \in I \) has a pullback in $C$. Let \( P \) be the
pullback object, with morphisms: 
\begin{itemize}[itemsep=-2pt]
\item \( p_1: P \to A \),
\item \( p_2: P \to B \),
\end{itemize}
satisfying \( i \circ p_1 = j \circ p_2 \), and the universal property
as described above. We need to show that \( p_1 \) and \( p_2 \) are
in $I$. 

\paragraph{Show \( p_1: P \to A \) is in $I$.}
Consider the pullback square:
\[
\xymatrix{
  P \ar[r]^{p_2} \ar[d]_{p_1} & B \ar[d]^j \\
  A \ar[r]_i & C
}
\]
Here, the morphism \( j: B \to C \) is in $I$. By the \emph{stability
  under pullbacks} property of the inclusion system, if the morphism
\( j: B \to C \) (the right vertical leg) is in $I$, then the opposite
leg \( p_1: P \to A \) (the left vertical leg) is also in $I$. Thus: 
\[
p_1 \in I.
\]

\paragraph{Show \( p_2: P \to B \) is in $I$.}
Now consider the same pullback square, but focus on the other leg:
\[
\xymatrix{
  P \ar[r]^{p_2} \ar[d]_{p_1} & B \ar[d]^j \\
  A \ar[r]_i & C
}
\]
Here, the morphism \( i: A \to C \) (the bottom horizontal leg) is in
$I$. Again, by the stability under pullbacks property, if \( i: A \to
C \) is in $I$, then the opposite leg \( p_2: P \to B \) (the top
horizontal leg) is in $I$. Thus: 
\[
p_2 \in I.
\]

\paragraph{Conclusion of Existence.}
Since the co-span has a pullback \( P \), and we’ve shown that both \(
p_1: P \to A \) and \( p_2: P \to B \) are in $I$ using the stability
property, the pullback \( P \) is indeed a pullback “consisting of
abstract inclusions.” 

\paragraph{Address Uniqueness.}
In category theory, pullbacks are unique up to unique
isomorphism. Suppose \( P' \) is another pullback of the co-span, with
morphisms: 
\begin{itemize}[itemsep=-2pt]
\item \( p_1': P' \to A \),
\item \( p_2': P' \to B \),
\end{itemize}
such that \( i \circ p_1' = j \circ p_2' \), and \( P' \) satisfies
the universal property. There exists a unique isomorphism \( \phi: P'
\to P \) such that: 
\[
p_1' = p_1 \circ \phi, \quad p_2' = p_2 \circ \phi.
\]
We need to ensure that \( p_1', p_2' \in I \). Since \( P' \)
is also a pullback of the same co-span, applying the same argument: 
\begin{itemize}[itemsep=-2pt]
\item \( j: B \to C \in I \) implies \( p_1': P' \to A \in
  I \), 
\item \( i: A \to C \in I \) implies \( p_2': P' \to B \in
  I \). 
\end{itemize}
Alternatively, consider the compositions:
\begin{itemize}[itemsep=-2pt]
\item \( p_1' = p_1 \circ \phi \), where \( p_1 \in I \) and
  \( \phi \) is an isomorphism, 
\item \( p_2' = p_2 \circ \phi \), where \( p_2 \in I \) and
  \( \phi \) is an isomorphism. 
\end{itemize}
In many inclusion systems, $I$ is stable under composition with
isomorphisms (e.g., in sets, composing an injection with a bijection
yields an injection). If this property holds, \( p_1', p_2' \in
I \). However, the stability under pullbacks alone is
sufficient, as applied directly to \( P' \), so we don’t need to
assume additional closure properties. 

Thus, any pullback of the co-span has projection morphisms in $I$, and
all such pullbacks are isomorphic via unique isomorphisms preserving
the diagram. 

\paragraph{Step 4: Finalize the Proof.}
Since the co-span has a pullback, that pullback has projection
morphisms in $I$ due to the stability of $I$ under pullbacks. The
uniqueness follows from the categorical property of pullbacks: any two
pullbacks are uniquely isomorphic, and all satisfy the same property
regarding $I$. 

Thus, in a category $C$ with an inclusion system $I$, for any co-span
of abstract inclusions \( A \xrightarrow{i} C \xleftarrow{j} B \)
(with \( i, j \in I \)) that has a pullback, there exists a
pullback with projection morphisms in $I$, and it is unique up to
isomorphism, satisfying the problem’s requirement. 

The statement is proved.

% ------------------------------------------------------------------------
\end{document}